# Solving integer multi-objective optimization problems using TOPSIS, Differential Evolution and Tabu Search

Renato A. Krohling
Department of Production Engineering &
Graduate Program in Computer Science, PPGI UFES
Federal University of Espírito Santo
Av. Fernando Ferrari, 514, CEP 29075-910 Vitória,
Espírito Santo, ES, Brazil
krohling.renato@gmail.com

Erick R. F. A. Schneider
Former Graduate Student
PPGI UFES- Federal University of Espírito Santo Av.
Fernando Ferrari, 514, CEP 29075-910 Vitória, Espírito
Santo, ES, Brazil
erickrfas@gmail.com

## ABSTRACT

This paper presents a method to solve non-linear integer multi-objective optimization problems. First the problem is formulated using the Technique for Order Preference by Similarity to Ideal Solution (TOPSIS). Next, the Differential Evolution (DE) algorithm in its three versions (standard DE, DE best and DEGL) are used as optimizer. Since the solutions found by the DE algorithms are continuous, the Tabu Search (TS) algorithm is employed to find integer solutions during the optimization process. Experimental results show the effectiveness of the proposed method.

## 1. INTRODUCTION

In general, Genetic Algorithms - GA [13] converges to a single final solution when used for optimization purposes. Schaffer [19] was the first to develop GA for solving multi-objective optimization (MOO) problems based on ideas proposed by Rosenberg [18] called Vector Evaluated Genetic Algorithm - VEGA. Srinivas and Deb [21] proposed a non-dominated sorting Genetic Algorithm - NSGA to solve MOO problems, which have been improved by Deb et al. [10] and known as NSGA-II, and largely used for MOO problems with continuous variables. Zitzler &Thiele [29] proposed the Strength Pareto Evolutionary Algorithm- SPEA to solve MOO problems with a quantitative analysis comparing many existing algorithms in the literature. The SPEA algorithm uses cluster analysis to reduce the size of the Pareto set. It also employs a technique to maintain the diversity of the solutions. Zitzler et al. [28] proposed also an improved version of SPEA, known as SPEA-II algorithm.

Zhang and Li [25] presented a MOEA/D algorithm based on decompositions. The MOO problem is transformed into scalar sub-problems, which are optimized simultaneously. Each sub-problem is optimized using information from its several neighboring sub-problems. The algorithm presents a lower computational complexity than the NSG-II algorithm.

Zhang and Cui [24] presented a Particle Swarm Optimization - PSO to solve MOO problems combined with the Technique for order Preference by Similarity to Ideal Solution - TOPSIS [23] to choose solutions within the Pareto set.Azzam and Mousa [1] proposed a method based on GA and the concept of dominance to solve the MOO problem of reactive power compensation, which was modeled as an optimization problem with three conflicting objectives. This approach also uses the TOPSIS.Zheng et al. [26] developed a hybrid approach basedon GA to solve the component-based product design problem that was formulated as a MOO problem. This hybrid algorithm includes a strategy to maintain the diversity of the population and a clustering method.There are few methods in the literature for non-linear integer MOO problems. Bellahcene [2] proposes a method that requires that the objective functions be monotone. Vassilev and Narula [16] propose a method requiring that the functions be linear. Genova and Guliashki [11] developed a method where the constraints must be convex functions.

In the last few years, Differential Evolution - DE [22]has been successfully applied to multimodal problems with promising results to find a single solution of optimization problems[6]. However, for integer multi-objective problems ithas not been applied yet, as far as we know.The DE algorithm provides a good balance between exploration and exploitation during the search improving the performance of the optimization process.In this work, the DE algorithm is used as optimizer to explore the continuous search space, but since DE has no memory, this motivated us to combine the algorithm with Tabu Search -TS [12]. In this paper, we propose a method which uses only the value of the objective function and the constraints, without no assumptions regarding linearity, continuity, and convexity of the objective functions and the constraints. The proposed method consists of three stages, where each stage uses the DE algorithm hybridized with the Technique for Order Preference by Similarity to Ideal Solution - TOPSIS[15] to transform a $d$-objective optimization problem in single-objective optimization problems. In the last stage, hybridization with TSalgorithm to find integer solutions from the real solutions obtained previously is performed.

The remainder of this paper is organized as follows. In Section 2, we present the mathematical formulation of the optimization problem. The three algorithms used in our approach are described in Section 3. In Section 4, we present a hybrid method to find solutions of integer multiple-objective optimization problems. Computational results for three benchmarks are provided in Section 5 to show the suitability of the method. In Section 6, we give some conclusions with directions for further works.

## 2. Problem formulation

A usual notation is used here: $\mathbb{Z}$ denotes the set of integers, $\mathbb{R}$ the set of real numbers, and $\mathbb{Z}^n = \mathbb{Z} \times \overset{n}{\cdots} \times \mathbb{Z}$. The problem of interest in this paper is the non-linear integer multi-objectiveoptimization problem, which, without loss of generality, can be defined as:

$$\min(f_1(x), f_2(x), ..., f_i(x), ..., f_d(x))$$
subject to: (1)
$$g_j(x) \leq 0, j = 1, ..., m \quad \text{and } x \in \mathbb{Z}^n.$$

where $f_i : \mathbb{Z}^n \to \mathbb{R}$ are the $d$ objective functions, $g_j : \mathbb{Z}^n \to \mathbb{R}$ are inequality constraints; $l_j$ and $u_j$ represents the lower and upper bounds of the integer search space, respectively.

Next, some definitions for MOO problems are presented:

**Definition 1.1 (Dominance):** a solution $a$ dominates another solution $b$, i.e, $a \prec b$, if occur simultaneously

$$f_1(a) \leq f_1(b) \text{ and ... and } f_d(a) \leq f_d(b) \quad (2)$$
$$f_1(a) < f_1(b) \text{ or ... or } f_d(a) < f_d(b) \quad (3)$$

**Definition 1.2 (Pareto Solution):** a candidate solution is a Pareto solution if there are not in the set of feasible solutions of the problem another candidate solution that dominates it [8].

**Definition 1.3 (Pareto Front):** The Pareto front of a MOO problem is the curve formed by the image of the application of the $d$ objective functions of the problem on the set of Pareto solutions [8].

The interested reader is referred to [4] for more information on MOO. Next, we describe each algorithm separately, and then in the next section the proposed method using a hybrid approach for solving integer MOO problems is presented.

## 3. Algorithms description

A method to solve multi-objective optimization problems is developed, whereas the TOPSIS is used to formulate the multi-objective problem in mono-objective problems. Next, the DE and TOPSIS are used to solve the resulting mono-objective problems. At the last stage of the proposed method, the TS algorithm is executed to find integer solutions provided by DE algorithm. In the following, we describe the three algorithms used.

### 3.1 The technique for order preference by similarity to ideal solution

TOPSIS [14] is a technique to find the best alternative(s) for a multi-criteria decision making (MCDM) problem consisting of alternatives and criteria (*benefit* or *cost*). The algorithm selects the best alternative within the set of alternatives so that there is a compromise among the criteria. The algorithm calculates the positive ideal solution (PIS) and the negative ideal solution (NIS) and chooses the alternative that best approximates the PIS and at the same time increase the distance to the NIS. The PIS is defined as the solution containing the best values of the criteria within the set of alternatives, whereas the NIS contains the worst values within the set of alternatives.

$$A = \begin{pmatrix} f_{1,1} & \cdots & f_{1,d} & g_1 \\ f_{i,1} & \cdots & f_{i,d} & g_i \\ f_{NP,1} & \cdots & f_{NP,d} & g_{NP} \end{pmatrix} \quad (4)$$

First, we define a matrix $A = (x_{i,j})_{NP \times d+1}$ which contains in each line an alternative. In this paper, an alternative means a possible solution. The matrix $A$ is composed of $d+1$ columns (criteria). The first $d$ criteria are the objective function for each candidate solution (or alternative), while the last criterion is the maximum value of the constraint violation.

In our notation $f_{i,j}$ means the evaluation of the alternative $i$ ($i = 1, ..., NP$) evaluated according the objective $j$ ($j = 1, ... d+1$) and constraints are treated as an additional objective function as $g_i = \max(0, g_1(\overline{X_i}), ..., g_m(\overline{X_m}))$.

The weight vector composed by individual weights $w_j (j = 1, ..., d+1)$ to each criteria $C_j$ satisfies $\sum_{j=1}^{d+1} w_j = 1$. In this paper, the matrix $A$ is normalized for each criterion $C_j$ through

$$p_{ij} = \frac{x_{ij}}{MAX \ \overline{x}_j}, \text{ with } j = 1, ..., d+1 \text{ and } MAX \ \overline{x}_j \quad \text{represents}$$

the maximum value for each criterion $C_j$. Thus, the decision matrix $B$ represents the normalized relative rating of alternatives and is described by:

$$B = (p_{i,j})_{NP \times (d+1)} \quad (5)$$

The TOPSIS [14] begins with the calculation of the positive ideal solutions $B^+$ (benefits) and negative ideal solutions $B^-$ (cost) as follows:

$$B^+ = (p_1^+, p_2^+) \text{ where } p_j^+ = (\max_i p_{ij}, j \in J_1; \min_i p_{ij}, j \in J_2) \quad (6)$$

$$B^- = (p_1^-, p_2^-) \text{ where } p_j^- = (\min_i p_{ij}, j \in J_1; \max_i p_{ij}, j \in J_2) \quad (7)$$

where $J_1$ and $J_2$ represent the benefit and cost criteria, respectively.

Secondly, compute the Euclidean distances between $B_i$ and $B^+$ (benefits) and between $B_i$ and $B^-$ (cost) as follows:

$$d^+ = \sqrt{\sum_{j=1}^{d+1} w_j (d_{ij}^+)^2} \text{ where } d_{ij}^+ = p_j^+ - p_{ij}, \text{with } i = 1, ..., NP. \quad (8)$$

$$d^- = \sqrt{\sum_{j=1}^{d+1} w_j (d_{ij}^-)^2} \text{ where } d_{ij}^- = p_j^- - p_{ij}, \text{with } i = 1, ..., NP. \quad (9)$$

Next, one calculates the relative closeness coefficient $\xi_i$ to each alternative $B_i$ in relation to the positive ideal $B^+$ as:

$$\xi_i = \frac{d_i^-}{d_i^+ + d_i^-}. \quad (10)$$

Finally, ranking according to the relative closeness coefficient. The best alternatives are those that have the greatest value $\xi_i$ and should be chosen because they are closer to the positive ideal solution. In this paper, TOPSIS is used to select the best solution within a population of candidate solutions. In order to obtain a compromise among the set of criteria, which in this case are the value of the objective function and the maximum constraint violation. The TOPSIS pseudo-code is shown in Fig.1.

```
input: matrix A
B = normalization of input matrix A
Calculation of the PIS and NIS Solutions according to (6) and (7)
Calculation of Euclidean distances between $B_i$ and $B^+$ (benefits)
and between $B_i$ and $B^-$ (cost) according to (8) and (9),
respectively
Calculation of the relative closeness coefficient $\xi_i$ to each
alternative $B_i$ according to (10)
Ranking according to the relative closeness coefficient
Return the alternative with the greatest value of $\xi_i$
```

Fig.1: TOPSIS pseudo-code.

## 3.2 Differential evolution

The optimization algorithm Differential Evolution (DE) was introduced by Storn and Price [23]. Similar to other Evolutionary algorithms (EAs), DE is based on the idea of evolution of populations of possible candidate solutions, which undergoes the operations of mutation, crossover and selection [6]. The parameter vectors are denoted by with components $x_i$. The lower and upper bounds of $x_{i,j}$ are $l_j$ and $u_j$, respectively. The index represents the individual's index in the population and is the position in D-dimensional search space.

First, the population of candidate solutions is initialized as follows:

$$x_{i,j} = l_j + U(0,1)_{i,j}(u_j - l_j) \quad j=1,...n;\ i=1,...NP \quad (11)$$

where $U(0,1)_{i,j}$ is a uniformly distributed random number within [0, 1]. NP and $n$ stand for population size and problem dimension, respectively.

The DE algorithm undergoes the following operators: Mutation, Crossover and Selection.

Mutation is seen as a change or perturbation with a random element. In DE-literature, a parent vector from the current generation is called target vector, a mutant vector obtained through the differential mutation operation is known as donor vector and finally an offspring formed by recombining the donor with the target vector is called trial vector. The mutation operation of the standard DE is calculated as follows:

$$\vec{V}_i = \vec{X}_{r1,i} + F(\vec{X}_{r2,i} - \vec{X}_{r3,i}) \quad (12)$$

where $F$ is a scaling factor between 0.4 and 1, and $r_1, r_2, r_3 \in [1, NP]$, such that $r_1 \neq r_2 \neq r_3$.

The next variant of the DE studied is the $DE_{best}$. The mutation operation of the $DE_{best}$ is calculated as follows:

$$\vec{V}_i = \vec{X}_{r1,i} + F(\vec{X}_{r2,i} - \vec{X}_{gbest,i}) \quad (13)$$

where gbest is the best individual of the population.

The DEGL algorithm[20] differs from the standard DE due the inserting of a ring neighborhood topology as shown in Fig. 2. It favors a balance between exploration and exploitation improving the capabilities of DE algorithm. The donor vector is calculated by performing a linear combination of the local donor vector $\vec{L}_i$, created using members of a neighborhood, and between the global donor vector $\vec{G}_i$, created using members of the population. This linear combination is made by a weighting factor which weights exploration versus exploitation. The smaller the value of the weight, the higher is the exploration capacity of the algorithm. On the other side, the larger the weight, the higher is the exploitation capacity of the algorithm.

The mutation operator is calculated in three steps as follows:
1) Local donor vector creation.
2) Global donor vector creation.
3) Combination of local and global donor vectors.

The local donor vector is calculated as follows:

$$\vec{L}_i = \vec{X}_i + \alpha(\vec{X}_{best_i} - \vec{X}_i) + \beta(\vec{X}_{r1} - \vec{X}_{r2}) \quad (14)$$

where, $\vec{X}_{best_i}$ is the best neighbor of $\vec{X}_i$, and the neighborhood is defined by a ring topology.

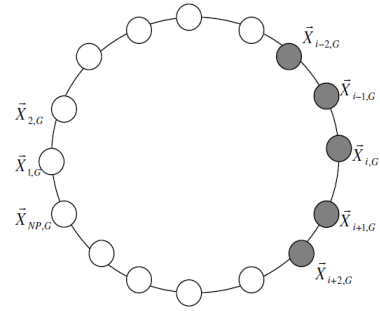

Figure 2. Ring Topology.

The neighborhood size is defined by a parameter $k$, where in the case of Fig. 2, it was set to $k = 2$. $\vec{X}_i$ is the $i$-th individual of the neighborhood, $r_1$ and $r_2$ are chosen within the neighborhood so that $i \neq r_1 \neq r_2$.

The global donor vector $\vec{G}_i$ is calculated as follows:

$$\vec{G}_i = \vec{X}_i + \alpha(\vec{X}_{gbest} - \vec{X}_i) + \beta(\vec{X}_{q1} - \vec{X}_{q2}) \quad (15)$$

where $\vec{X}_{gbest}$ is the best individual in the population, $q_1$ and $q_2$ are chosen in the population such that $i \neq q_1 \neq q_2$.

The donor vector $\vec{V}_i$ is calculated as a linear combination of $\vec{L}_i$ and $\vec{G}_i$ as:

$$\vec{V}_i = r\vec{G}_i + (1-r)\vec{L}_i \quad (16)$$

where $r$ is a weighting factor that weights exploration versus exploitation and is calculated as:

$$r = \frac{number\_of\_iterations}{max\_iterations}. \quad (17)$$

The operation of crossover and selection are the same for all variants of the DE.

In crossover step, a recombination is generated between the vector $\vec{X}_i$ and the vector $\vec{V}_i$ as follows:

$$u_{i,j} = \begin{cases} v_{i,j} & \text{if } U(0,1) \leq C_r \text{ or } j = j_{rand} \\ x_{i,j} & \text{otherwise.} \end{cases} \quad (18)$$

where $C_r$ is the crossover rate, and $j_{rand}$ is a uniformly distributed random value between 1 and $n$ to ensure that at least one component of $\vec{V}_i$ is part of the vector $\vec{U}_i$.

The selection phase consists in the choice of individuals from the current population to the next, while keeping as members of the population the best individuals. Each individual $i$ in the population $\vec{X}$ is compared with its respective individual $i$ of the population $\vec{U}$ as:

$$\vec{X}_{i,G+1} = \begin{cases} \vec{U}_{i,G} & \text{if } \vec{U}_{i,G} \text{ is better than } \vec{X}_{i,G} \\ \vec{X}_{i,G} & \text{otherwise.} \end{cases} \quad (19)$$

---

**Input** population size $NP$, crossover probability $C_r, \alpha$ and $\beta$.

**for** i = 1 to $NP$
    **for** j = 1 to $n$
        $x_{i,j} = x_{j,\min} + U(0,1)_{i,j}(x_{j,\max} + x_{j,\min})$
    **end for**
**end for**
**while** termination condition not met
    **for** i = 1 to NP
        choose $p$ and $q$ such that $i \neq p \neq q$ from $[i-k, i+k]$
    $\vec{L}_i = \vec{X}_i + \alpha(\vec{X}_{best_i} - \vec{X}_i) + \beta(\vec{X}_p - \vec{X}_q)$
        choose $p$ and $q$ such that $i \neq p \neq q$ from $[1, NP]$
    $\vec{G}_i = \vec{X}_i + \alpha(\vec{X}_{gbest} - \vec{X}_i) + \beta(\vec{X}_p - \vec{X}_q)$
    $\vec{V}_i = r\vec{G}_i + (1-r)\vec{L}_i$
    $j_{rand} = U(1,n)$
        **for** j = 1 to $n$
            **if** $U(0,1) \leq C_r$ **or** j = $j_{rand}$
    $u_{i,j} = v_{i,j}$
            **else**
    $u_{i,j} = x_{i,j}$
            **endif**
        **end for**
        **if** $f(\vec{U}_i) \leq f(\vec{X}_i)$
    $\vec{X}_i = \vec{U}_i$
        **end if**
    **end for**
**end while**
**return** $\vec{X}$

**Figure 3. DEGL pseudo-code.**

---

An individual is considered better than another [8]:

An individual is considered better than another [8]:

- If both individuals are feasible, it is chosen the individual with better fitness.
- If one individual is feasible and the other infeasible, it is chosen the feasible individual.
- If both individuals are infeasible, it is chosen the one that least violates the constraints.

The pseudo-code of DEGL is presented in Fig. 3 for unconstrained optimization problems as similarly described in [5]. In our case, the constraint handling is done using TOPSIS (DEGL and DE$_{best}$) and Deb [8].

### 3.3 Tabu Search

The Tabu search, proposed by Glover[12], is alocal search algorithm which uses a structure of memory to performmovements allowing the algorithm to escape from local optima. This structure of memory, called tabu list, stores the most recent movements. Then, thealgorithm examines the neighbors of a current point, and selects the best neighbor nottaboo (not on the list), storing the previous move in the Tabu list, whichacts as a queue of fixed size, i.e., the first in is the first out.This process is repeated until a maximum number of iterations is reached. Theoperation of the algorithm depends on a starting point from which itperforms movements. The closer the starting point is from the optimum, the faster the convergence of the algorithm to this optimum is.

Since the solutions found in the previous stage are real values, firstly they are converted to integer by means of rounding as described inpseudo-code of Fig.4. Next, the rounded solutions are used as initial solutions for the Tabu Search algorithm as described in pseudo-code ofFig.5 [17].

A solution is considered better than another in the same manner as in the selection operation of DE algorithm.

The Tabu Search algorithm calls the Tabu Move routine in its inner loop as presented in Fig. 6.In order to test the feasibility of the candidate integer solutions, it is used the same procedure for constraint handling as previously described by [8].

---

**Input** vector $\vec{X}_i$

**for** j = 1 to n
    **if** $\left(U(0.1) \leq x_{i,j} - \lfloor x_{i,j} \rfloor\right)$
$x_{i,j} = \lceil x_{i,j} \rceil$
    **else**
$x_{i,j} = \lfloor x_{i,j} \rfloor$
    **end if**
**end for**
**return** $\vec{X}_i$

**Figure 4.Rounding of real to integer solutions.**

---

The arguments of this algorithm are the current solution $\vec{X}_i$, the best solution found $\vec{X}_i^*$, the current iteration $j$, and the tabu vector $t$. The Tabu information is kept in the vector $t$ and$t_c$ holds the iteration at which thevariable $c$ has been updated.

The beginning of the algorithm prevents the block of the search in which all the moves are potentially taboo. Then, a random movement is selected and a value within its limits is chosen for the movement. The neighborhood is searched by means of a breadth first search [17]. With breadth first all the neighbor solutions are checked, and the best is returned.

```
Input: solution X̄ᵢ
X̄ᵢ = Construct(X̄ᵢ)
X̄* = X̄ᵢ
t = (−n,...,−n)
for j = 1 to N
    X̄ᵢ = TabuMove(X̄ᵢ, X̄*, j, t)
        if X̄ᵢ is better than X̄*
    X̄* = X̄ᵢ
    end if
end for
return X̄*
```

**Figure 5. Tabu Search Algorithm.**

```
Input: solution X̄ᵢ, best solution X̄*, current iteration k and tabu list t
if k − tⱼ > n  ∀j
    c = U(1,n)
    xᵢ,c = U(lc, uc)
    tc = k
        return X̄ᵢ
end if
V̄ = X̄ᵢ
for j = 1 to n
    S̄ = X̄ᵢ
    d = U(1,n)
        for each δ ∈ {−1,1}
    sⱼ = xᵢ,ⱼ + δ
            if sⱼ ∈ [lⱼ, uⱼ] and S̄ better than V̄
                if k − tⱼ > d or S̄ better than X̄*
    V̄ = S̄
                c = j
                end if
            end if
        end for
end for
X̄ᵢ = V̄
tc = k
returnX̄ᵢ
```

**Figure 6. Tabu Move Algorithm.**

## 4. The hybrid method

The proposed method to solve multi-objective problems consists in three stages, where in each stage is used the DE+TOPSIS to solve mono-objective optimization problems. The DEGL used is similar to that presented in [5]. However, for the choice of local or global best solution, the TOPSIS is used, whereas each candidate solution corresponds to an alternative, and each objective function corresponds to a criterion, which is considered either a *benefit or cost*. *Benefit* and *cost* criteria are related to maximizing and minimizing the objective function, respectively. Since the problem consists in minimizing the objective functions, then both criteria are considered as *cost*. Thus, the objective function and the constraints are the criteria, so the TOPSIS find a compromise solution minimizing both criteria.

The multi-objective optimization problem is given as

$$\min(f_1(x), f_2(x),..., f_i(x),..., f_{k-1}(x))$$
$$\text{subject to } g_i(x) \leq 0 \text{ with } i = 1,...,m$$
$$\text{with } x \in X \quad (20)$$
$$\text{and } l_j \leq x_j \leq u_j \text{ with } j = 1,...,n$$

The first stage [15] consists in finding the solutions PIS and NIS of the problem to be optimized. The PIS and NIS are defined as follows:

$$\text{PIS: } (\min f_1(x),..., \min f_k(x)) = \left(f_1^*,...,f_k^*\right) \quad (21a)$$

$$\text{NIS: } (\max f_1(x),..., \max f_k(x)) = \left(f_1^-,...,f_k^-\right) \quad (21b)$$

$x \in X$.

For the second stage, two functions are defined as follows [14]:

$$d^{NIS}(x) = \left\{\sum_{j=1}^{k} w_j^2 \left[\frac{f_j^- - f_j(x)}{f_j^- - f_j^*}\right]^2\right\}^{1/2} \quad (22a)$$

$$d^{PIS}(x) = \left\{\sum_{j=1}^{k} w_j^2 \left[\frac{f_j(x) - f_j^*}{f_j^- - f_j^*}\right]^2\right\}^{1/2} \quad (22b)$$

where $w_j$ is the weight of the $j$-th objective function. In this paper, we use equal weights in this stage. Next, DEGL+TOPSIS algorithm is used to solve the problems (22a) and (22b).

The goal is to minimize $d^{PIS}(x)$ and to maximize $d^{NIS}(x)$. To this end a compromise solution is sought. The compromise solution is found using membership functions. However, before finding the compromise solution, $\left(d^{PIS}\right)^*$ and $\left(d^{NIS}\right)^*$ are calculated using the DEGL+TOPSIS as follows:

$$\left(d^{PIS}\right)^* = \min d^{PIS}(x), \text{ such that } x \in X, \text{ with solution } x^p$$

$$\left(d^{NIS}\right)^* = \max d^{NIS}(X), \text{ such that } x \in X, \text{ with solution } x^n$$

$$\left(d^{PIS}\right)' = d^{PIS}(x^n) \text{ and } \left(d^{NIS}\right)' = d^{NIS}(x^p).$$

The membership functions are described as:

$$\mu_1(x) = \begin{cases} 1 & \text{if } d^{PIS}(x) < (d^{PIS})^* \\ 1 - \dfrac{d^{PIS}(x) - (d^{PIS})^*}{(d^{PIS})' - (d^{PIS})^*} & \text{if } (d^{PIS})' \geq d^{PIS}(x) \geq (d^{PIS})^* \\ 0 & \text{if } d^{PIS}(x) > (d^{PIS})'. \end{cases} \quad (23a)$$

$$\mu_2(x) = \begin{cases} 1 & \text{if } d^{NIS}(x) > (d^{NIS})^* \\ 1 - \dfrac{(d^{NIS})^* - d^{NIS}(x)}{(d^{NIS})^* - (d^{NIS})'} & \text{if } (d^{NIS})' \leq d^{NIS}(x) \leq (d^{NIS})^* \\ 0 & \text{if } d^{NIS}(x) < (d^{NIS})'. \end{cases} \quad (23b)$$

To minimize (23a) and maximize (23b), we use the max-min operator proposed by Bellman and Zadeh [3] and extended by Zimmermann [27]. So, the problem is reformulated as:

$$\mu_D(x^*) = \max\{\min(\mu_1(x), \mu_2(x))\}. \quad (24)$$

The max-min solution is illustrated in Fig.7.

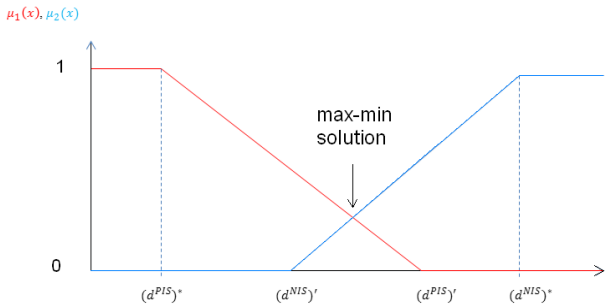

**Figure 7. Compromise solution obtained by means of membership functions.**

This problem is then reformulated as follows:

$$\max \alpha$$
subject to:
$$\mu_1(x) \geq \alpha \quad (25)$$
$$\mu_2(x) \geq \alpha$$
with $x \in X$.

At this point, one begins the third stage, which consists of solving the problem (25). We use the algorithm DE+TOPSIS setting the α-level of satisfaction in the objective functions and constraints in the last objective function. We alternate the DE+TOPSIS algorithm and the TS algorithm. After a fixed number of iterations of the DE+TOPSIS, the TS algorithm is executed to convert the problem of real to integer and to perform a local search near of the solutions found by the DE+TOPSIS algorithm. Next, DE+TOPSIS is executed again accompanied by the TS, i.e., the algorithms are alternated to solve the problem.

It is known that the solution of the problem (25) is also the solution of the problem (20) [15]. So, at the end of the third stage is provided a solution of the problem (20). The proposed approach inspired by our previous work [20] is shown in Fig.8.

To solve the problem (1), the constraints are inserted as an additional objective and the proposed method is applied to solve the resulting multi-objective problem.

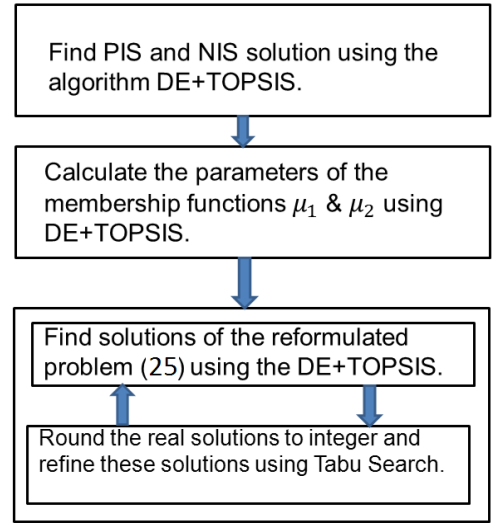

**Figure 8. Illustration of the proposed approach.**

## 5. Computational results

In this paper, we use three non-linear integer multi-objective optimization problems to evaluate the proposed method.

Problem 1 [2]:

$$\max f_1(x) = 2x_1 + 5x_2$$
$$\max f_2(x) = 3x_1x_2 - x_1 + 6x_2$$
$$\max f_3(x) = 2x_1^2 + x_1x_2 - x_2$$

subject to:

$$x_1 + 2x_2 + 2.9\sqrt{0.09x_1^2 + 0.05x_2^2 + 1} \leq 18$$
$$3x_1 + 2x_2 \leq 22.$$
$$1 \leq x_1 \leq 7$$
$$1 \leq x_2 \leq 5$$
$$x_1, x_2 \in \mathbb{Z}.$$

The known solutions [2] are: (4,4), (2,5), (6,2) e (5,3).

Problem 2 [11]:

$$\min f_1(x) = x_1^2 + 3x_2^2$$
$$\min f_2(x) = 5x_1^2 + x_2^2$$
$$\min f_3(x) = 2x_1^2 - x_2$$

subject to:

$$-x_1 - x_2 + 11 \leq 0$$
$$0 \leq x_1, x_2 \leq 16$$
$$x_1, x_2 \in \mathbb{Z}.$$

The known solutions [11] are (1,10), (2,9), (3,8), (4,7), (5,6), (6,5), (7,4), (8,3), (9,2) e (10,1). However, the solutions (7,4) e (8,3) dominate the solution (10,1), and the solution (8,3) dominates (9,2).

Problema 3 [16]:

max $f_1(x) = x_1$

max $f_2(x) = x_2$

subject to:

$2x_1 - x_2 \leq 21$

$5x_1 + 1.5x_2 \leq 57.5$

$4x_1 + 5x_2 \leq 61.1$

$6x_1 + 15x_2 \leq 135$

$x_2 \leq 6.5$

$x_1, x_2 \in \mathbb{Z}$.

The only known solution to this problem is (9,5) as given by Narula and Vassilev [16].

The parameter setup for the algorithms DE (standard), $DE_{best}$ and DEGL used in the experiments is given in Table 1.The value of DE parameters was suggested by [5, 6], who studied this issue in-depth. It is out of scope of this paper to investigate the optimal parameters of DE. Obviously, we have carried out preliminary experiments to our particular problems investigated here in order to adopt such DE parameters as listed in the Table 1.The TOPSIS algorithm does not have any parameters, and we adopt the same weights for all criteria. The parameters of the TS are the number of executions, and the list size is equal to the problem dimension (number of variables).The weighting factor $r$ is as given by (17).

Table 1: Parameter setup for the algorithms used in the experiments.

| Population size | 40 |
|---|---|
| Number of iterations of DE | 100 |
| Number of iterations of the TS | 1000 |
| Number of alternations between DE and TS | 10 |
| Crossover rate | 0.9 |
| $\alpha$ Parameter | 0.8 |
| $\beta$ Parameter | 0.8 |
| Scaling Factor $F$ | 0.8 |
| Neighborhood size ($k$) | 2 |

Table 2: Success rate for the problem 1.

| Solutions found to Problem 1: | DEGL | $DE_{best}$ | DE |
|---|---|---|---|
| (2,5) | 18 (90%) | 14(70%) | **20(100%)** |
| (4,4) | **20 (100%)** | **20(100%)** | **20(100%)** |
| (5,3) | **20 (100%)** | **20(100%)** | **20(100%)** |
| (6,2) | **20 (100%)** | **20(100%)** | **20(100%)** |

Table 3: Success rate for the problem 2.

| Solutions found to Problem 2: | DEGL | $DE_{best}$ | DE |
|---|---|---|---|
| (6,5) | 16 (80%) | 7(35%) | **17(85%)** |
| (2,9) | 15 (75%) | 11(55%) | **17(85%)** |
| (4,7) | 15 (75%) | 11(55%) | **18(90%)** |
| (5,6) | 14 (70%) | 12(60%) | **19(95%)** |
| (3,8) | 17 (85%) | 12(60%) | **20(100%)** |
| (7,4) | 15 (75%) | 10(50%) | **19(95%)** |
| (8,3) | 15 (75%) | 6(30%) | **19(95%)** |
| (0,11) | 11(55%) | 7(35%) | **18(90%)** |
| (0,14) | 10(50%) | 11(55%) | **19(95%)** |
| (1,10) | **15(75%)** | 9(45%) | 14(70%) |
| (0,15) | 10(50%) | 12(60%) | **19(95%)** |
| (0,12) | 9(45%) | 9(45%) | **18(90%)** |
| (0,16) | 10(50%) | 18(90%) | **20(100%)** |
| (0,13) | 12(60%) | 9(45%) | **19(95%)** |

Table 4: Success rate for the problem 3.

| Solutions found to Problem 3: | DEGL | $DE_{best}$ | DE |
|---|---|---|---|
| (9,5) | **20 (100%)** | 12(60%) | 7(35%) |
| (10,4) | **19 (95%)** | 12(60%) | 11(55%) |
| (11,1) | 17 (85%) | **20(100%)** | 19(95%) |
| (7,6) | **19 (95%)** | 9(45%) | 15(75%) |
| (5,7) | 8(40%) | 11(55%) | **18(90%)** |

The hybrid algorithm has been applied to the benchmark problems 1 2, and 3 and executed 20 times.The solutions obtained during the optimization process and the success rate are shown in Table 2. Numbers in bold mean the best value obtained. In the Tables 2, 3 and 4 we can see the effectiveness of DE in finding multiple solutions. It is observed that the standard DE, apparently showed better results than $DE_{best}$ and DEGL, but further in-depth studies need to be performed with additional benchmarks.

## 6. Conclusion

In this paper, we propose a novel method for solving non-linear integer multi-objective optimization problems. This method has the advantage that only the value of the objective function and the constraints areneeded. The effectiveness of the DE versions investigated depends on the problem at hand. Preliminary results are promising but further studies are necessary in order to indicate what EA is more appropriate for this kind of challenging problem. We intend to extend the approach to find solutions of non-linear integer bi-level optimization problems.

## 7. Acknowledgments

The first author would like to thank the financial support of the Brazilian agency CNPq.